\documentclass[11pt,a4paper]{article}
\usepackage[hyperref]{emnlp-ijcnlp-2019}
\usepackage{times}
\usepackage{latexsym}

\usepackage{url}
\aclfinalcopy % Uncomment this line for the final submission

\usepackage{verbatim}

\usepackage{graphicx}

\title{Unsupervised Natural Question Answering with a Small Model}

\author{Martin Andrews \\
  Red Dragon AI  \\
  Singapore \\
  {\tt martin@reddragon.ai} \\\And
  Sam Witteveen \\
  Red Dragon AI  \\
  Singapore \\
  {\tt sam@reddragon.ai} \\}

\date{}

\begin{document}
\maketitle
\begin{abstract}
%This document provides a basic paper template and submission guidelines.
%Abstracts must be a single paragraph, ideally between 4--6 sentences long.
%Gross violations will trigger corrections at the camera-ready phase.

%Fact extraction from text is an important endeavour, but the 
The
recent demonstration of the power of huge language models 
such as GPT-2 to memorise the answers to factoid questions
raises questions about the extent to which knowledge 
is being embedded directly within these large models.
%
%The recent demonstration of the power of huge language models, 
%specifically that the GPT-2 had memorised 
%the answers to questions in the Natural Questions dataset,
%raises the question of whether training such large models 
%is essential to making progress in Natural Language Understanding.
%
%Moreover : Are ever-larger models required to learn facts about 
%
This short paper describes an architecture through which much smaller
models can also answer such questions - 
by making use of `raw' external knowledge.
The contribution of this work is that 
%In contrast to other work, 
the methods presented here rely on unsupervised learning techniques, 
complementing the unsupervised training of the Language Model.
%even when trained in a fully unsupervised manner.
%
The goal of this line of research is to be able to 
add knowledge explicitly, without extensive training.  

\end{abstract}

\begin{comment}
\newpage

\section{Extra Text}
very small numbers of supervised training examples, and make use of unlabelled data.
\end{comment}

\section{Introduction}

The field of question answering has been dominated by supervised methods 
for competitive tasks such as the Stanford question answering dataset
(SQuAD) \cite{rajpurkar2016squad}.  
However, as discussed in \citet{yogatama2019learning},
some of these datasets are becoming over-optimised for, 
making the architectures less generally applicable.

%In the arena of more general architectures, 

At the other extreme, 
the ability of the GPT-2 \cite{radford2019language} model to answer factoid questions, 
based purely on unsupervised training directed 
at improving its Language Model (LM) performance, was striking.
But further reflection highlights the following issues :
\begin{itemize}
\item Questions correctly (and confidently) answered were a small fraction ($\sim$1\%) of the questions asked
\item Huge model size and long training periods were required before such behaviour was manifested
\item This does not appear to be a practical approach to adsorbing an extensive knowledgebase
\end{itemize}

This work describes early work in aiding generalised models such as GPT-2
to answer questions, without having to embed 
%the 
facts directly in the model's weights.  
The overall direction of work is towards encouraging such generalised
models to make use of external datasources (and other resources) 
without having to internalise all the data in models of exponentially 
increasing size (e.g. GPT-2-1.5B is more than 10x the size of GPT-2-117M).

\section{Natural Questions Dataset}

The Natural Questions (NQ) dataset \cite{natural_questions_47761} 
is a question answering dataset containing 
307,373 training examples, 7,830 development examples, and 7,842 test examples. 
Each example is comprised of a google.com query and a corresponding Wikipedia page. 
Each Wikipedia page has a passage (or long answer) 
annotated on the page that answers the question 
and one or more short spans from the annotated passage 
containing the actual answer. 
The long and the short answer annotations can however be empty. 
If they are both empty, then there is no answer on the page at all. 
If the long answer annotation is non-empty, 
but the short answer annotation is empty, 
then the annotated passage answers the question 
but no explicit short answer could be found. 
Finally, 1\% of the documents have a passage annotated with a short
answer that is `yes' or `no', instead of a list of short spans.

%The recently introduced Natural Questions dataset 
%\cite{natural_questions_47761}
%is a promising resource to test this more quantitatively. 

As reported in \citet{radford2019language}, 
GPT-2-1.5B answers 4.1\% of NQ questions correctly when evaluated 
by the exact match metric commonly used on reading
comprehension datasets like SQuAD.  
In contrast, the smallest GPT-2-117M model 
(used as the basis for the model proposed in this work) 
is reported as not being capable of exceeding the 1.0\% accuracy of 
the simple baseline which returns the most
common answer for each question type (who, what, where, etc...). 
The fact that GPT-2-1.5B answered 5.3 times more questions correctly
suggests that model capacity has been a major factor in
the poor performance of neural systems on this kind of task
as of yet. 

%The 30 most confident answers generated by GPT-2 
%on development set questions are shown in Table 5. 

\section{Model Architecture}

The model proposed here is built from several components which
include (a) 876k Wikipedia sentences, 
addressible via embeddings;
(b) a pretrained GPT-2-117M language model which was noted to be incapable of 
answering questions successfully in \citet{radford2019language}; and
(c) a scheme for incorporating `sentence hints' 
into the language generation context.

\subsection{Embeddings for Sentence Lookup}

Three different 
%sentence 
embedding methods were used : 

(i) pre-trained BERT-base (L=12, H=768, A=12, Total Parameters=110M) \cite{devlin2018bert}, 
using the the $\mathtt{bert-as-service}$ Python tool\footnote{\url{https://bert-as-service.readthedocs.io/}}.
For a given input sentence this returns a 768-d embedding, 
calculated as the GlobalAveragePooling of the top-but-one 
layer of the pretrained BERT model;

(ii) Smooth Inverse Frequency (SIF) \cite{arora2017asimple} embeddings, 
calculated by inverse-frequency weighting the BPE embeddings
(from the GPE-2-117M model being used for the text generation task)
followed by removal of the first PCA component; and 

(iii) Universal Sentence Encoder \cite{cer2018universal}, 
the training details not clear in the 
%unclear in 
paper, but USE is not a purely unsupervised model : 
``We augment unsupervised learning with training on supervised data 
from the Stanford Natural Language Inference (SNLI) corpus'' \cite{snli:emnlp2015}.

Methods (i) and (ii) were not fine-tuned on the question answering task 
(since this would violate the spirit of this unsupervised-only system), 
whereas method (iii) was included to judge the benefits of adding some supervised
training to the embedding stage.

\begin{table*}[th]
\caption{Sample question answers with filter examples, 
and examples of answers where pure SQuAD accuracy did not make sense 
when the base data included far more information than the original (single) 
wiki article targetted by the Natural Questions dataset.}
\label{qanda-table}
\vskip 0.15in
\begin{center}
\begin{small}
%\begin{sc}

\begin{tabular}{llll}
\hline
Question & Target & GPT-2-117M & Reject reason \\
\hline
Who is the richest club in the championship?  &
`Aston Villa', &
The richest club in  &
{\sc Smart Alec}
\\
& `Manchester City' & the championship & \\
\hline
Are all firestone tires made in the usa? & 
`NO' & 
No & 
{\sc Y/N question}
\\
\hline
What is the name of manchester united stadium? & 
`Old Trafford' & 
Manchester United & 
{\sc Within question}
\\
\hline
Who cracked the enigma code in world war 2? & 
`Turing' & 
Alan Turing & 
{\sc N/a : Accepted}
\\
\hline
How many inches is the iphone 5s screen?	&
`4 - inch screen size', & 
4 inches & 
{\sc N/a : Accepted}
\\
&`4 in', `4 in ( 10 cm )' &  & 
\\
\hline
\end{tabular}

%\end{sc}
\end{small}
\end{center}
\vskip -0.1in
\end{table*}

\subsection{Embeddings for Questions}

In order that facts might be supplied by external text, 
embeddings $e(s_n)$ were produced 
for each sentence $s_n$ of the $N(=876,645)$ wikitext sentences, 
and also $e(q_j)$ was calculated for each $q_j$ of the $J$ questions.

The search term was calculated by adding a `question to sentence' vector, 
set to the mean difference between the embeddings for question phrases 
and those of wikitext sentences to the original question $q_j$  : 
\[
search_j = e(q_j) + \frac{1}{N}\sum{e(s_\cdot)} - \frac{1}{J}\sum{e(q_\cdot)}
\]

%\subsection{LM Question Answering}
%A pretrained GPT-2.117M model was used to produce the answer text.
%The prompt technique illustrated in Table 1 was used, 
%since it proved more effective when combined with the `hint' mechanism described below.
%Explain importance of `preamable' choice

\subsection{Knowledge Look-up}

In order to aid the LM in retrieving factoid answers, 
`hint sentences' sufficient to fill half of the LM context window 
were retrieved
%\footnote{The $\mathtt{annoy}$ Python package was found to be very efficient, see \url{https://github.com/spotify/annoy}}
from the list of the $N$ wikitext sentences, using
%a Manhattan distance metric ranking of the $s_n$ vs $search_j$
a cosine distance ranking of the $s_n$ vs $search_j$

\begin{figure}[t]
  \vskip 0.0in
  \begin{center}
    \centerline{\includegraphics[width=\columnwidth]{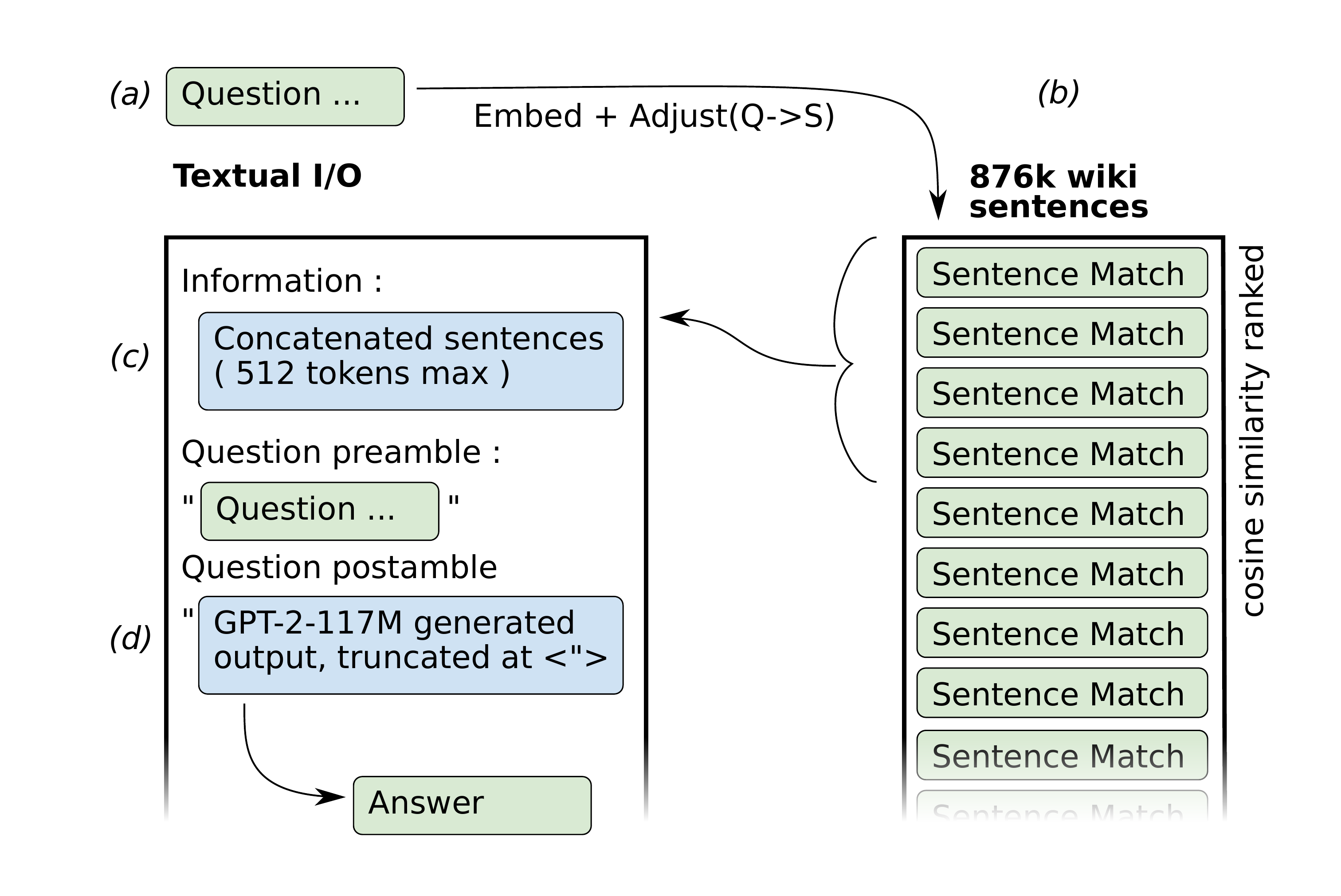}}
    \caption{
    Proposed information flow : (a) Initial question; (b) Wiki sentence ranking; (c) hinting in preamble; (d) GPT2 output.
    }
    \label{model-diagram}
  \end{center}
  \vskip -0.2in
\end{figure}

\subsection{LM Context Seeding}

In order to obtain the results in \citet{radford2019language} 
for the NQ task, their GPT-2-1.5B model context was seeded with 
example question/answer pairs which helped the model 
infer the short answer style of the dataset.

Rather than expect the smaller GPT model to extrapolate from 
the Q \& A format,
%directly
both the `hint sentences' and the question $q_i$ 
were incorporated into the context seen by the model directly:

\begin{quote}
Information :

$\mathtt{HintSentence{[}~{]}}$ {\it or} $\mathtt{None}$

The best short answer to ``$q_i$?'' from the 
information above is `` \ldots
\end{quote}

The GPT-2-117M output is then recorded up until the closing double-quote
(closing quotes appears to be strongly favoured by the LM).

%When the hint sentences was used, it was simply inserted in the lines 
%above the question preamble, 
%so that it becomes the `information above' sought by the LM.

\subsection{Sampling from the Language Model}

A number of approaches to sampling from the model were tried
(including Beam search, which performed poorly), 
and the following were found to work satisfactorially :

\begin{enumerate}
\item  SoftMax temperature was kept at 1.0 (i.e. as trained);

\item  Nucleus Sampling \cite{holtzman2019curious} was used, 
with only tokens that cover the first 90\% of probability space
being considered as choices at each step.  This appears to give 
a good mix of diversity without `going off the rails' - which is
desirable for human-like communication \cite{grice1975logic};

\item   A probability bias term \cite{murray2018correcting} was added to the 
log-probabilities of each sequence, whereby each token was `awarded' a bonus of $\alpha$, 
which was found empirically to create a more balanced spread of long and short outputs;

\item   After a sorted list of 100 different sequences was created, this was 
further filtered (as illustrated in Table \ref{qanda-table}) to reject answers that were very unlikely to be correct: 

     \begin{itemize}
\item answers that simply repeat the question (determined as whether 
the answer's bigram Jaccard similarity with the question exceeds 0.5);
\item answers that are contained within the question verbatim;
\item answers such as `yes/no', `i don't know', `none', `no one', `it depends' - 
which may have been safe choices, 
but could not score positively on the filtered list of questions.
     \end{itemize}
\end{enumerate}

Further details can be found in the Supplimental Materials.
%Further details can be found in the model source code, 
%to be made available upon publication.

\section{Experiments}

The model architecture was applied to the NQ task, 
and results are reported for performance on the validation set
(the training set was unused).  
Only questions that were (a) not Yes/No; and (b) had a `short answer' were considered, 
resulting in 3975 triples of \{question, wikitext, answer list\}.

The list of `hint sentence' candidates was set to be 
the aggregate of all the sentences across the 3975 wikitext pages,
totalling $\sim$876k sentences.  
Importantly, the hint sentence choices weren't restricted to 
the wikitext corresponding to the specific question - 
which makes the task significantly more difficult that the 
BERT baseline for Natural Questions task \cite{alberti2019bert}, which 
works on an article-by-article basis.

%The performance of GPT-2 is still 
%significantly worse than the 30 to 50\%
%range of open domain question answering systems which
%hybridize information retrieval with extractive document
%question answering \cite{alberti2019bert}.

%and 
%(ii) using the Universal Sentence Encoder of \cite{cer2018universal} 
%(trained to match sentence
%

%All the sentences from all the NQ validiation examples 
%were used as the pool of `hint sentences' 
%(i.e. the hint sentences could come 
%from any of the  within whole 3975
%wikitexts - 

In the results reported, to reduce noise, the `Yes/No' questions 
were removed from consideration
(since scoring positively on these examples may the result of a coin-flip).

\section{Results}

This work is in its early stages, and the results obtained so far are encouraging,
despite being low in number.

For the 3975 useful NQ development set questions, 
we found that the poor results of using GPT-2-117M unaided reported 
in \citet{radford2019language} were born out.  
%The `No Hints' setting resulted in 0.84\% correct responses.

However, when using each question to select `hint sentences' 
from the whole list of 876k wikitext sentences, 
the GPT-2-117M was able to make use of the extra information 
(without having been explicitly training to do so).

\begin{table}[h]
\caption{Question answering accuracy.}
\label{results-table}
\vskip 0.15in
%\vskip 0.05in
\begin{center}
\begin{small}
\begin{sc}

%\begin{tabular}{lcccr}
%\toprule
%Data set & Naive & Flexible & Better? \\
%\midrule
%Vehicle   & 44.9$\pm$ 0.6& 61.5$\pm$ 0.4& $\surd$ \\
%\bottomrule
%\end{tabular}

\begin{tabular}{lrrr}
\hline
Embedding & dim & $\alpha$ & Score \\
\hline
No Hints  & - & 0.0   &  0.84\% \\
\hline
BERT-REST & 768 & 0.0   &  1.08\% \\
SIF       & 768 & 0.7   &  3.14\% \\
SIF       & 768 & 0.2   &  3.29\% \\
\hline
USE       & 512 & 0.0   &  4.45\% \\
\hline
\end{tabular}

\end{sc}
\end{small}
\end{center}
\vskip -0.1in
\end{table}

Note that the results in Table \ref{results-table} are not directly comparable with the 
reported accuracy of the 1.5 billion parameter GPT-2-1.5B (4.1\%), 
since the ``Yes/No'' questions have been deliberately excluded in the experimental results above,  
since random chance would then add approximately 1.8\% (of pure noise) 
to the results presented here.  Adjusting the reported GPT-2 figures 
(downward) for this effect shows that the proposed model has higher performance
for a much lower parameter count, even when using purely unsupervised training methods.

% base : 0.66%  WithYN : 2.79
% USE  : 4.45%  WithYN : 6.14

\section{Discussion}

%As mentioned in the Introduction 
As mentioned in \citet{sutskever-openai-video}, 
an online video in which \citet{radford2017learning} is discussed,
`higher order' capabilities seem to appear in language-related models
only if the size of the model is sufficient to have captured 
many the basic features of the underlying language,
since knowing the basic words and structures is more important to a Language Modeling 
objective than higher order features like sentiment and story arc (for instance).

Being able to capture such higher order features 
provides a natural incentive to want to scale the training of language models
to as large a number of parameters as possible.  
And undoubtedly there will be important and interesting results to come out of these efforts.

However, it is not at all clear that embedding factoids 
in neural network weights is a practical way of building intelligent systems.
Even humans (built on a biological neural substrate) 
seem to reason about facts symbolically \textit{despite} the processing being 
based in neurons.

The goal of this research is to explore how to interface 
the extremely effective aspects of models such as GPT-2 
with more accessible sources of knowledge and planning.  

By using the \textit{human readable} output of a Language Model 
component to direct further information gathering (or, potentially, other activities), 
one might imagine the system would not only become more capable (without exponentially long training),
but would also have an \textit{internal dialogue} that would be human interpretable.

\subsection{Further Work}

Clearly, more experimentation is needed to understand how to improve the current
system.  Fortunately, that can be accomplished without a huge investment in hardware.

In terms of sentence embedding techniques, 
one additional method was investigated, so far without encouraging results : 
the generation of sentence embeddings from 
using an additional layer for the GPT-2-117M model
it its initially untrained state.  This deserves further work, 
given the findings of \citet{wieting2019training}.

Also interesting is the potential for training a more specific retrieval/utilisation 
engine in a supervised manner, such as in \citet{Bapna_2019}, and then expanding the domain across which retrieval
is performed to encompass a much broader range of accessible facts without further training the model.
However, this is slightly contrary to the goal herein of using purely unsupervised techniques.

Beyond these initial phases, though, there is the potential for the
system to achieve some level of self-improvement. 
As was discussed in \citet{radford2019language}, the GPT-2-1.5B model
could not only answer some factoid questions, but it also had a 
good (self-) model of confidence in its answers\footnote{
``The probability GPT-2 assigns to its generated
answers is well calibrated and GPT-2 has an accuracy of
63.1\% on the 1\% of questions it is most confident in.''}.
This implies that if a trainable embedding component 
were included in \textit{this} paper's architecture it might be trainable 
(in a fully self-supervised way) to improve its self-hinting, 
and thereby achieve a self-improving positive feedback loop.

\section*{Acknowledgments}
The authors would like to thank Google for access to the TFRC TPU program which 
was used in training and fine-tuning models during experimentation for this paper.

%\section*{Acknowledgments}
%The authors would like to thank Google for access to the TFRC TPU program which 
%was used in training and fine-tuning models for this paper.
%The authors would like to thank Google for their support via credits on the Google Cloud Platform (GCP).

\bibliographystyle{acl_natbib}
\bibliography{../emnlp-ijcnlp-2019}

%\appendix

\end{document}